\documentclass{article} 
\usepackage{amsmath}
\usepackage{amsfonts}
\usepackage{amsthm}
\usepackage{dsfont}
\usepackage{epsfig}
\usepackage{upgreek}
\usepackage{graphicx}

\usepackage{setspace}

\usepackage{rotating}

\begin{document}

\title{{\bf A Canonical Image Set for Examining and\\ 
           Comparing Image Processing Algorithms}\vspace{0pt} }       
\author{
{\bf Jeffrey Uhlmann}\\
{\small Dept.\ of Electrical Engineering \& Computer Science}\\
{\small University of Missouri-Columbia}}
\date{}       
\maketitle

\begin{abstract}
The purpose of this paper is to introduce a set of four test images
containing features and structures that can facilitate effective 
examination and comparison of image processing algorithms. 
More specifically, the images are designed to more explicitly 
expose the characteristic properties of algorithms for 
image compression, virtual resolution adjustment, and
enhancement. This set was developed at the Naval Research 
Laboratory (NRL) in the late 1990s as a more rigorous alternative to 
Lena and other images that have come into common use for 
purely ad hoc reasons with little or no rigorous consideration of 
their suitability. The increasing number of test images 
appearing in the literature not only makes it more difficult to
compare results from different papers, it also introduces the
potential for cherry-picking to influence results.
The key contribution of this paper is the proposal to 
establish {\em some} canonical set to ensure that published 
results can be analyzed and compared in a rigorous way 
from one paper to another, and consideration of the four 
NRL images is proposed for this purpose.\\ 

\begin{footnotesize}
\noindent {\bf Keywords}: {\sf\footnotesize Test images, image compression, image superresolution, 
image enhancement, image processing, SVD.}
\end{footnotesize}
\end{abstract}

\section{Introduction}

This paper proposes a set of four specially-generated test images as 
candidates for general use in the qualitative assessment of image 
processing and related algorithms. The goal is to establish a consensus 
standard so that results from published experiments can be more 
easily and reliably compared. In other words, the objective is to
promote greater consistency in the assessment and presentation
of results in the literature. 

\begin{figure}
   \centering
   \includegraphics[width=\linewidth,keepaspectratio]{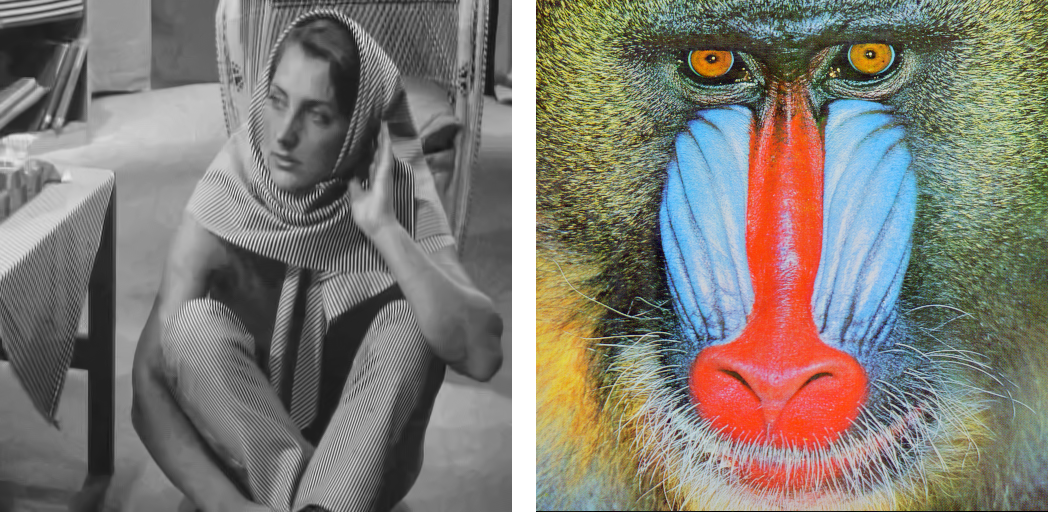}
\caption{Widely-used test images (a) {\em Barbara}  and  (b)  {\em Baboon}.}
\label{barboon}
\end{figure}

Presently there are many de facto standard test images
from which to choose when assessing a given image processing
algorithm. One example is the image {\em Barbara}, shown in
Fig.\,\ref{barboon}a. This image has very distinctive parallel line structures
that can be presumed useful for revealing clearly visible 
artifacts, e.g., in the form of moire patterns, after processing
by a given algorithm. The image {\em Baboon} (aka {\em Mandrill}), 
Fig.\,\ref{barboon}b, is
another widely-used test image which has appeal because of 
its distinctive mix of colors and textures. 

\begin{figure}
   \centering
   \includegraphics[width=0.49\linewidth,keepaspectratio]{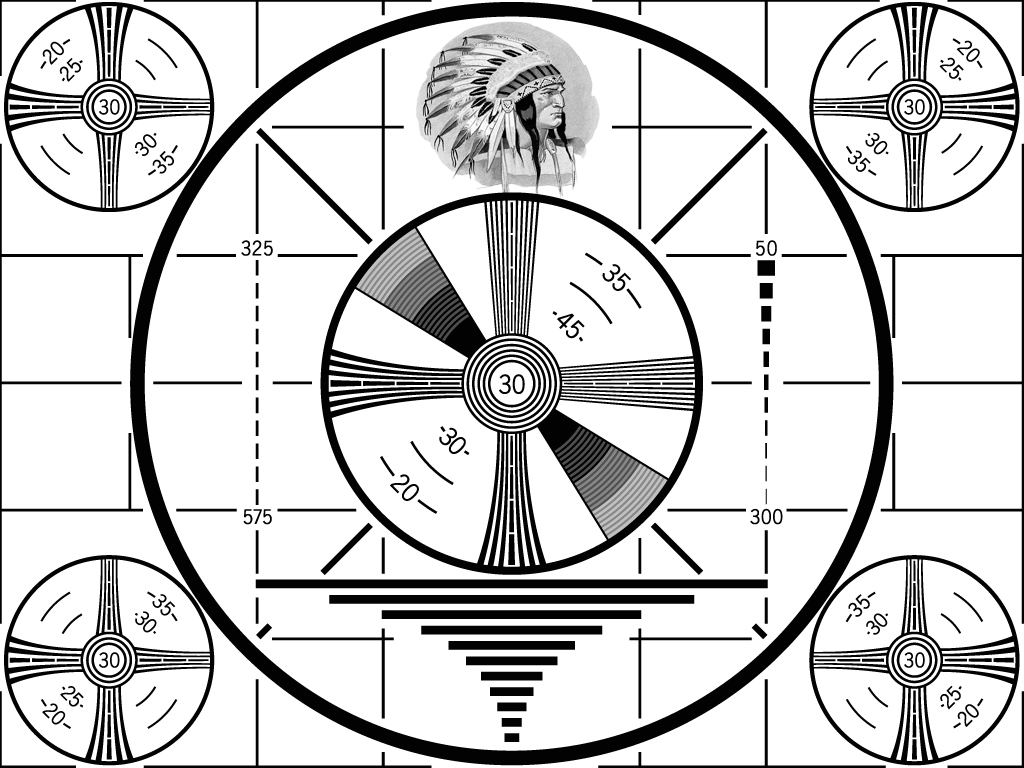}\vspace{3pt}
   \includegraphics[width=0.49\linewidth,keepaspectratio]{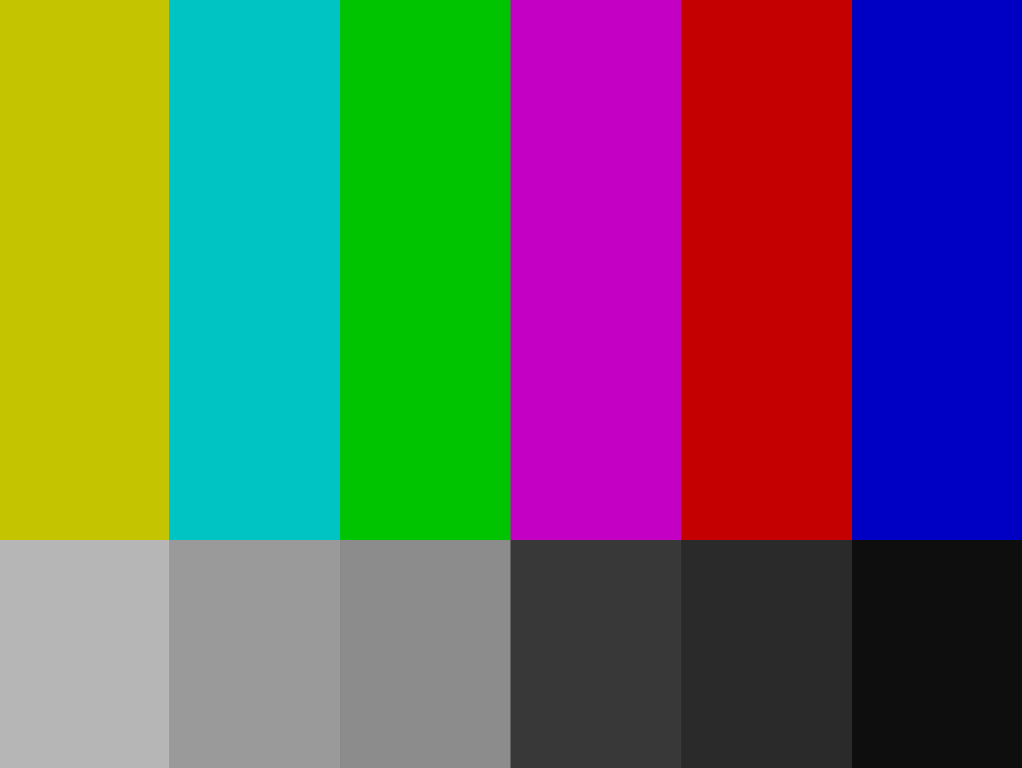}
\caption{Widely-used test images from the early days of television: (LEFT) {\em RCA Test Pattern}  
and  (RIGHT)  {\em Color Bars}.}
\label{tv}
\end{figure}

What is important to
note is that the assumed useful features of {\em Barbara}
and {\em Baboon} were determined {\em post hoc}, i.e., 
they were subjectively judged to have those features
rather than being intentionally produced to exhibit those
features in a rigorous form. Intentionally-produced test
images have been used in the past, especially in the
early days of television broadcasting. Fig.\,\ref{tv}a shows
an early RCA test pattern that was designed to reveal
artifacts of incorrect brightness and/or camera 
calibration~\cite{RCA}. 
Fig.\,\ref{tv}b shows color bars similar to those used later for
calibration of NTSC color television signals and color
rendering of computer monitors~\cite{Bars}.

The structure of this paper is as follows: Section~1 discusses
the difference between identifying the qualitative {\em properties}
of a given algorithm versus making a qualitative {\em assessment} of those
properties for a given application. Section~2 formally 
introduces the four proposed test images. Section~3
provides examples involving image super-resolution and
image compression. And Section~4 concludes with a 
brief summary and discussion.

\section{Assessing Image Processing Algorithms using Test Images}

Image processing algorithms can be assessed using purely
subjective qualitative judgments or by applying objective quantitative
formulas that may not accurately measure the salient properties
relevant to the intended application of interest. For example, suppose that
decompressed results from image compression algorithms $A$ and $B$
reveal that $A$ is superior in most cases to $B$ in terms of RMS error while
the soft blurring introduced by $B$ is judged in almost all cases to be 
aesthetically superior to the characteristic artifacts (e.g., ringing or 
blockiness) produced by $A$. Assuming that the algorithms offer
comparable compression ratios, which is ``better''? 

\begin{figure}
   \centering
   \includegraphics[width=2.05in,keepaspectratio]{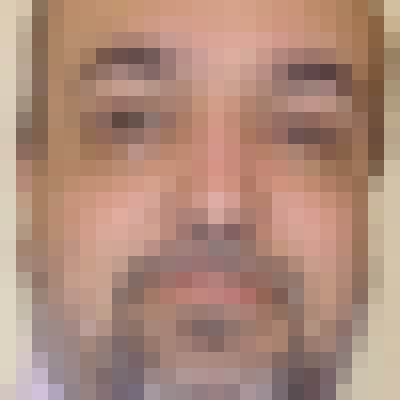}\vspace{3pt}
   \includegraphics[width=2.05in,keepaspectratio]{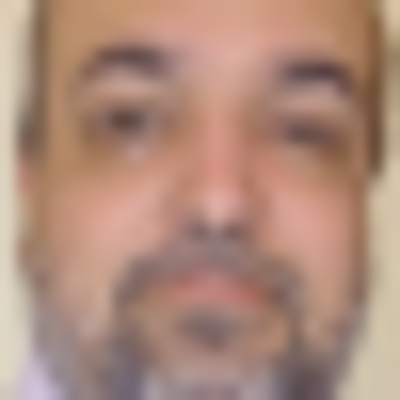}\vspace{3pt}
   \includegraphics[width=2.05in,keepaspectratio]{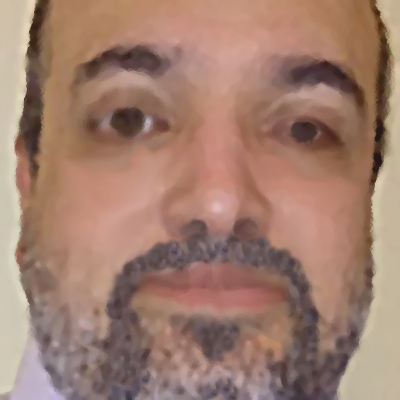}
\caption{Left-to-right shows ISR results from replication, cubic interpolation, 
and possible from machine learning.}
\label{marco}
\end{figure}

Image super-resolution (ISR) provides an illustrative example of 
the challenges that can arise when assessing competing approaches. 
ISR is an inherently ill-posed problem in that it requires the 
generation of intensity/color values for unobserved pixels,
so in some sense there is no ``correct'' or ``incorrect'' 
solution. Fig.\,\ref{marco} shows results for three methods for
synthetically increasing the resolution of a 
25x25-pixel image by a factor of 16.
The first method applies simple replication of pixels, the second
performs cubic polynomial interpolation between adjacent
pixels, and the third represents a result that might be generated
using a machine-learning algorithm based on a large database of
faces.

Pixel replication can be regarded as a very conservative 
approach to ISR that is unwilling to mix/combine intensities
of adjacent pixels to estimate intensities between them
whereas cubic interpolation assumes smooth intensity
gradations between pixels and generates unobserved
pixel intensities accordingly. One appealing property of
replication is that all information about the original
image is preserved, i.e., the original image can be 
recovered exactly. On the other hand, it introduces
uniform, high-frequency, rectilinear features that are 
purely artifacts of the algorithm in that they appear in 
all nontrivial images processed
by the method. Cubic interpolation, by contrast, 
introduces a blurring effect that in some sense is
suggestive of the fact that image detail is unavailable.

At first glance the machine-learning result in the example
of Fig.\,\ref{marco} may appear
``best'' in that it resembles an in-focus photograph of a
face, but much of the detail of that face is {\em not real} and 
thus may be misleading for applications in which the generated
image is intended to assist in the identification of the 
actual human subject of the original image. In a different application,
however, aesthetic considerations may be more important
than whether or not the synthetic face closely resembles
the original subject (it does not). The conclusion to be drawn is
that while  image processing algorithms can be assessed 
in terms of their characteristic properties, the qualitative 
ranking of different algorithms typically only makes sense with 
respect to application-specific criteria.

Over the years many images have become de~facto 
standards simply because they were used in papers 
that proved to be influential in the field. This is not 
unreasonable because subsequent researchers would naturally
be motivated to compare their alternative algorithms to the
prior state-of-the-art using the same images for comparison. 
In fact, if a
later paper were to use different images, questions
might be raised as to whether those images were selectively
chosen to yield more favorable results. Of course, similar
questions could also be raised about the choice of images
used in the original papers.

Attempts have been made to establish collections of ``standard'' test 
images (often including labels, e.g., {\em faces}~\cite{FERET}, 
{\em textures}~\cite{SIPI}, etc.), 
and these are valuable resources for deriving statistical measures of 
performance over many images that share a set of common features 
of interest. More specifically, they can be used to show that some
characteristic of a given algorithm that is observed when applied to
a few well-known standard images is robustly exhibited when applied
to a larger set of images. However, the availability of a large number of
images in a collection introduces opportunities for selection bias. What is 
needed is a small set of images that collectively captures the most critical
features of relevance to general image processing and thus may
be used as a common basis for comparing different algorithms 
across a range of application domains.

\section{Considerations for Choosing ``Standard'' Test Images}

\begin{figure}
   \centering
   \includegraphics[width=3.20in,keepaspectratio]{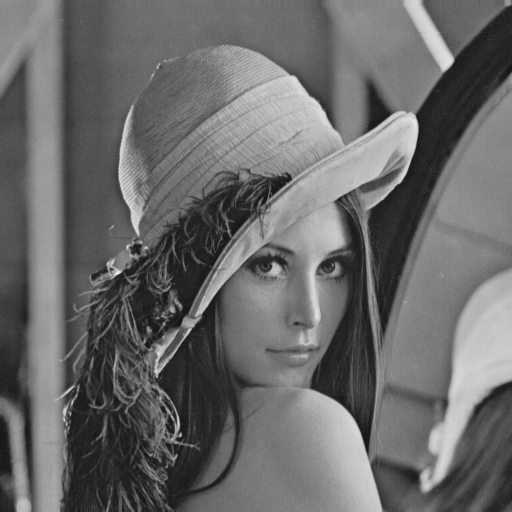}\vspace{3pt} 
   \includegraphics[width=3.20in,keepaspectratio]{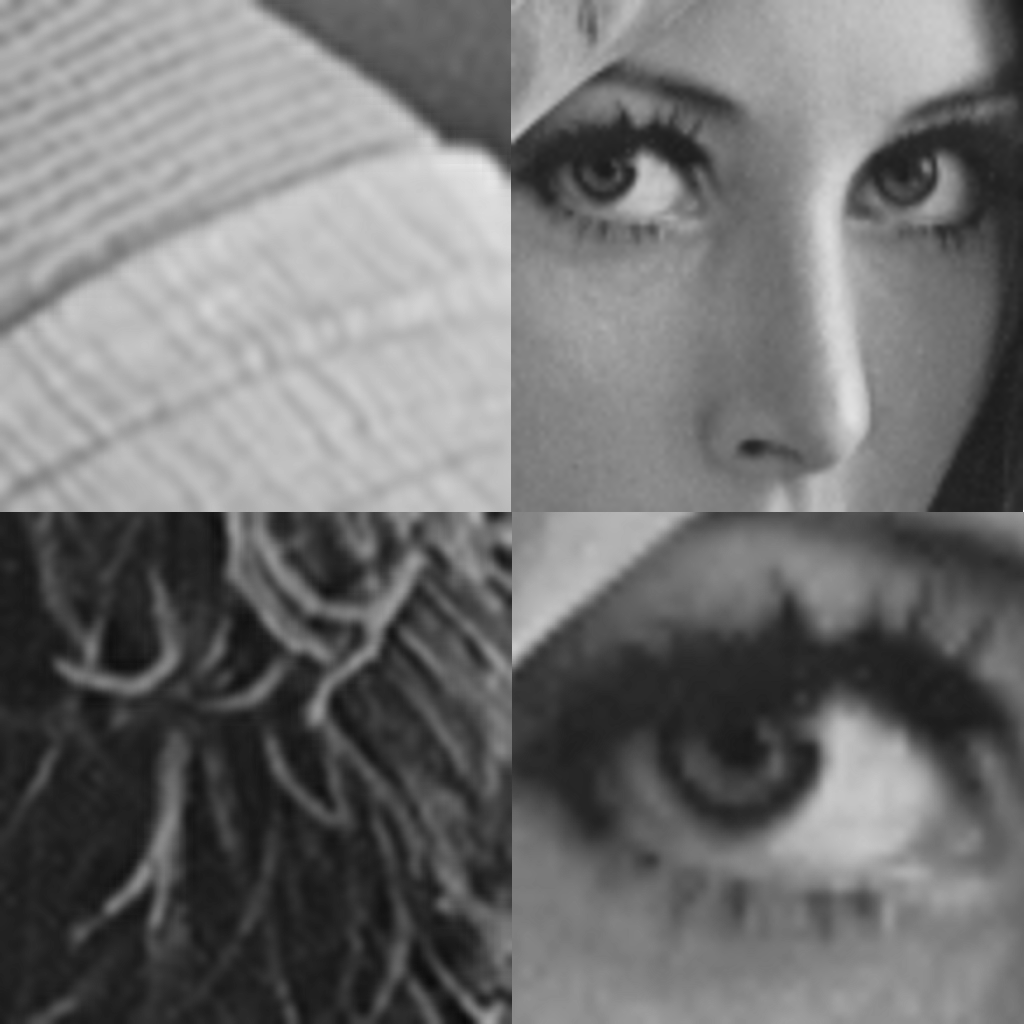}
\caption{{\em Left} is the standard Lena image and {\em Right}
                            are four commonly examined details from it.}
\label{lena}
\end{figure}

The ``{\em Lena}'' image (Fig.\,\ref{lena}) has been in widespread use for decades 
and is the most widely used standard in the field of image processing.
A major reason why {\em Lena} has been adopted so widely is because it
contains many of the feature types that are most commonly examined
when assessing image processing algorithms~\cite{Lena}. Specifically, 
the hat contains repetitive parallel weave structures; the feather 
contains complex textures; the skin of the face and shoulder show smooth
intensity gradations; and the eyes include familiar small-scale features
in the iris and lashes. Common undesirable artifacts generated from
compression or ISR processing of the {\em Lena} image include the 
appearance of checkerboarding in the weave of the hat; strong 
intensity discontinuities (banding) on the skin; and blurring of detail 
in the eyes. 

Instead of seeking out details (subregions) that exhibit features of
interest, e.g., edges, smooth intensity gradations, etc., a reasonable
question to ask is whether it might be preferable to construct a
small set of synthetic images, each element of which is tailored to 
clearly exhibit a particular class of feature, in place of a 
``real'' image such as {\em Lena}. Also, is it prudent to use 
images that include human faces, or text in a particular language,
when there is potential for evolutionary and/or cultural experience
to influence interpretation?  In the case of text, for example, 
strokes comprising characters in different languages often include 
ornamental flourishes (e.g., serifs) that native speakers may
unconsciously ignore, i.e., the relative attention given to
distinct textual features of similar size is culturally 
influenced. This obviously suggests that the 
assessed ``significance'' of artifacts introduced by 
algorithmic processing of text may be culturally biased.

More generally, the problem posed by most natural images
is that visual examination tends to be distracted by
interpretable content. For example, a compression
algorithm that identifies and attempts to preserve
approximately-straight lines may introduce significant
artifacts into the feather strands of Lena's hat that may be 
hardly noticeable under casual inspection if the more
easily-interpretable features of the face and shoulder
are rendered accurately. Of course this behavior of 
the algorithm could be an advantage in some applications,
but it is typically best to understand the properties of a
given algorithm first and then identify the applications
for which those properties are best suited. With this
in mind, four images are proposed as
candidates for standards for identifying and comparing
properties of image processing algorithms.

\section{Four Images}

The four images\footnote{Full-size images can be
obtained at: {\tt http://EntertainmentEngineering.missouri.edu/test-images.htm}} described in this section are 1024
pixels in height, 2048 pixels in width, with 256
greyscale intensities. The dimensions are chosen
to facilitate factor-of-2 decimations and to offer
typographical formatting flexibility in landscape
format, or vertical in multi-column format (as used
in this paper). Each 
image can also be subdivided into separate 
1024x1024 sub-images as needed. 

The first image, ``{\em Basic}'', consists of contiguous regions of
uniform intensity with boundaries of varying curvature. It
is suitable for revealing aliasing, pixelation, and ringing artifacts. 
\begin{figure*}
\centering
   \centering
   \includegraphics[width=\textwidth,keepaspectratio]{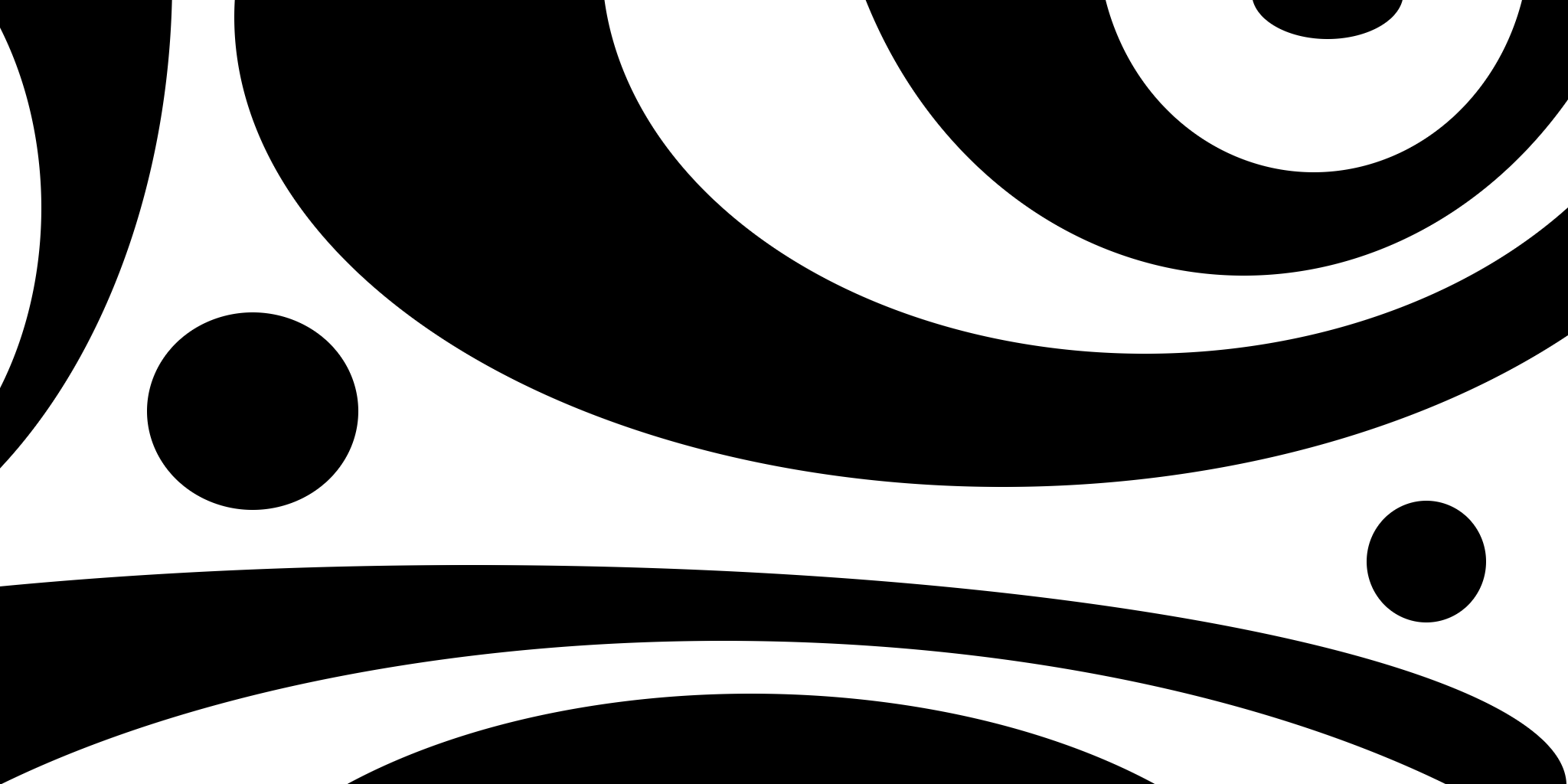}
\caption{``Basic'' - A simple set of contiguous regions of pure black and pure white.}
\label{basic}
\end{figure*}
As can be seen in Fig.\,\ref{basic}, {\em Basic} is a relatively low-information image and thus can be 
highly compressed. Because its information content is largely
focused along high-contrast curved edges, it is well-suited
for comparing compression algorithms at very high compression ratios
because artifacts can be expected to be most pronounced along
those edges. 

The second image, ``{\em Platonic}'', has smooth gradations of intensity
along contours of varying curvature. This image was constructed so that
intensity variations are at the limit for visual smoothness given the
available resolution and discrete intensity values. It can reveal,
for example, an algorithm's use of global contrast enhancement to
mask its blurring of edge detail. 
\begin{figure*}
\centering
   \includegraphics[width=\textwidth,keepaspectratio]{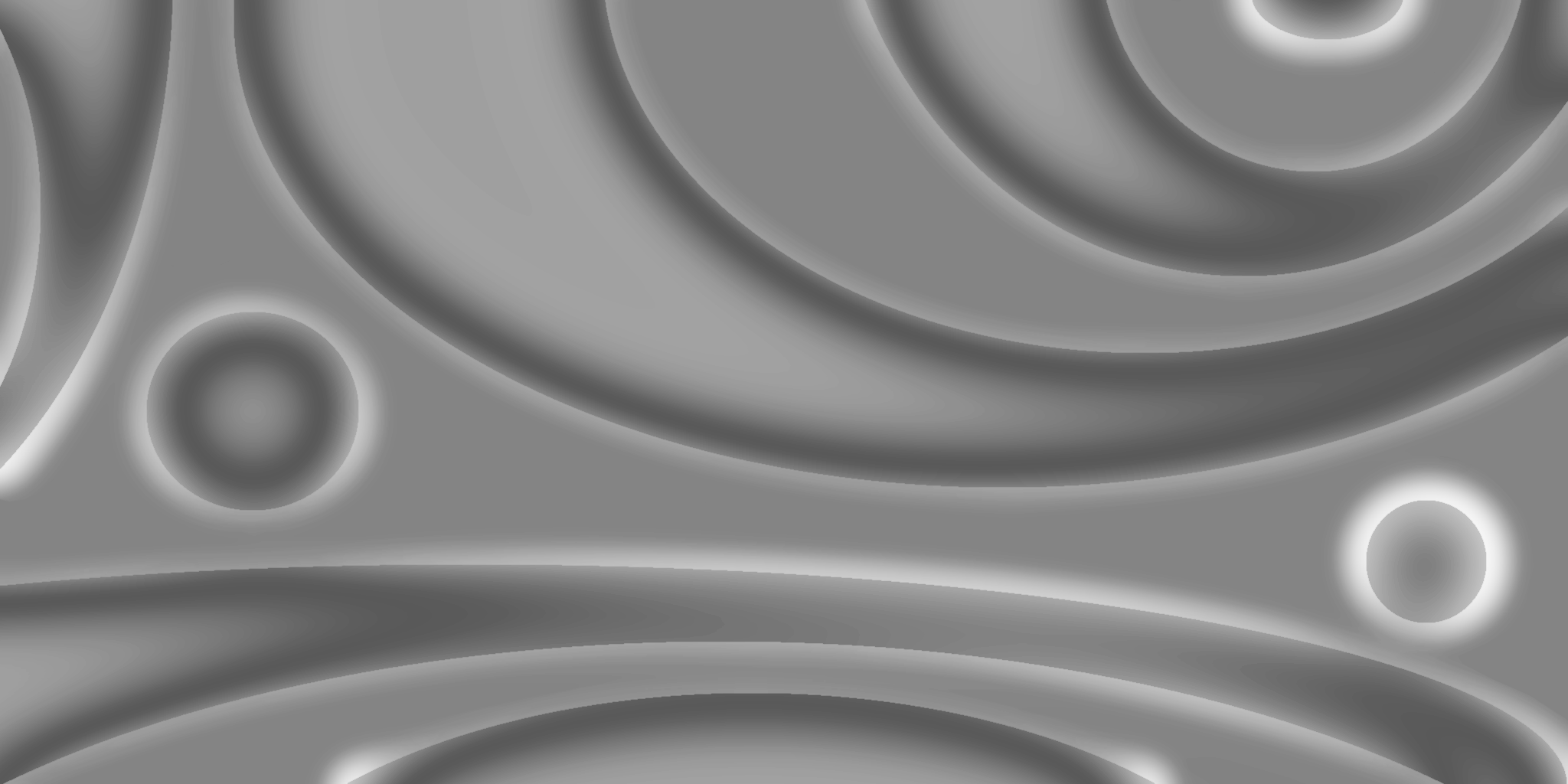}
\caption{``Platonic'' - Smooth intensity gradations.}
\label{platonic}
\end{figure*}
As can be seen in Fig.\,\ref{platonic}, {\em Platonic} contains a broad range of strong and weak intensity
gradients that vary with respect to boundary directions of curvature in 
such a way that can further enhance contrast and edge-enhancement 
artifacts. Interfaces between the bright diffusive regions around the 
circles and those of surrounding structures can be particularly
revealing of such artifacts.

\begin{figure*}
   \centering
   \includegraphics[width=\textwidth,keepaspectratio]{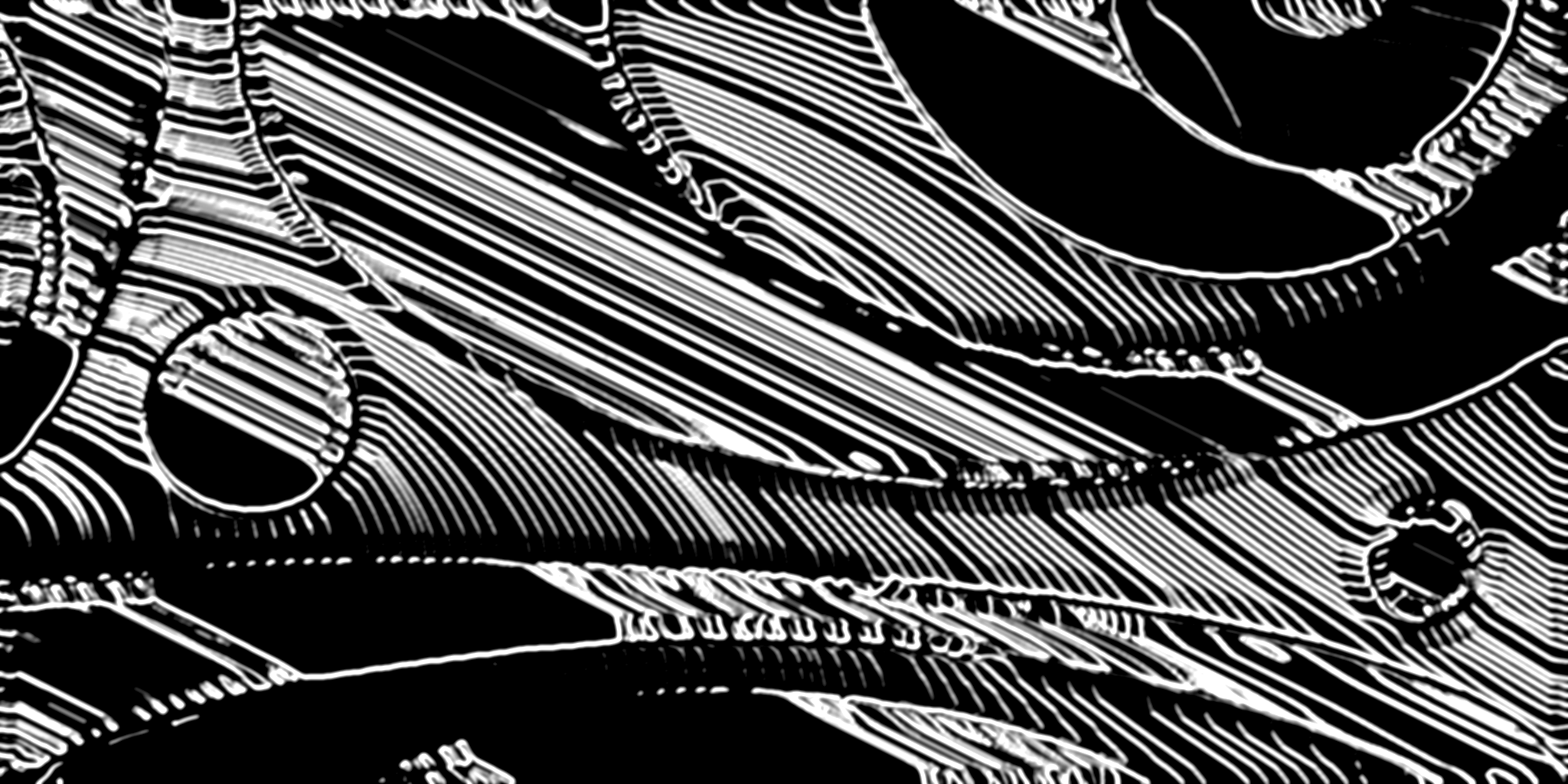}
\caption{``Natural'' - Structures common to natural images.}
\label{natural}
\end{figure*}

The third image, ``{\em Natural}'' (Fig.\,\ref{natural}), includes basic features found in natural images 
such as approximately-repeating structures with subtle variations in 
texture and illumination. This image is particularly tailored to reveal properties of 
an algorithm that are relevant to its use with radiological and multi-spectral 
imaging. Noise and defocusing artifacts can be introduced to this
image (or its intensity inverse) to model application-specific characteristics of an 
assumed sensing modality. 

As can be seen in Fig.\,\ref{natural}, 
{\em Natural} includes parallel structures of varying width, separation, and
curvature with smooth but nonlinear drop-offs in illumination. This image
is suitable for identifying loss of visibility or spurious presence of features
that may be diagnostically important, e.g., when interpreting X-ray
images.  

\begin{figure*}
   \centering
   \includegraphics[width=\textwidth,keepaspectratio]{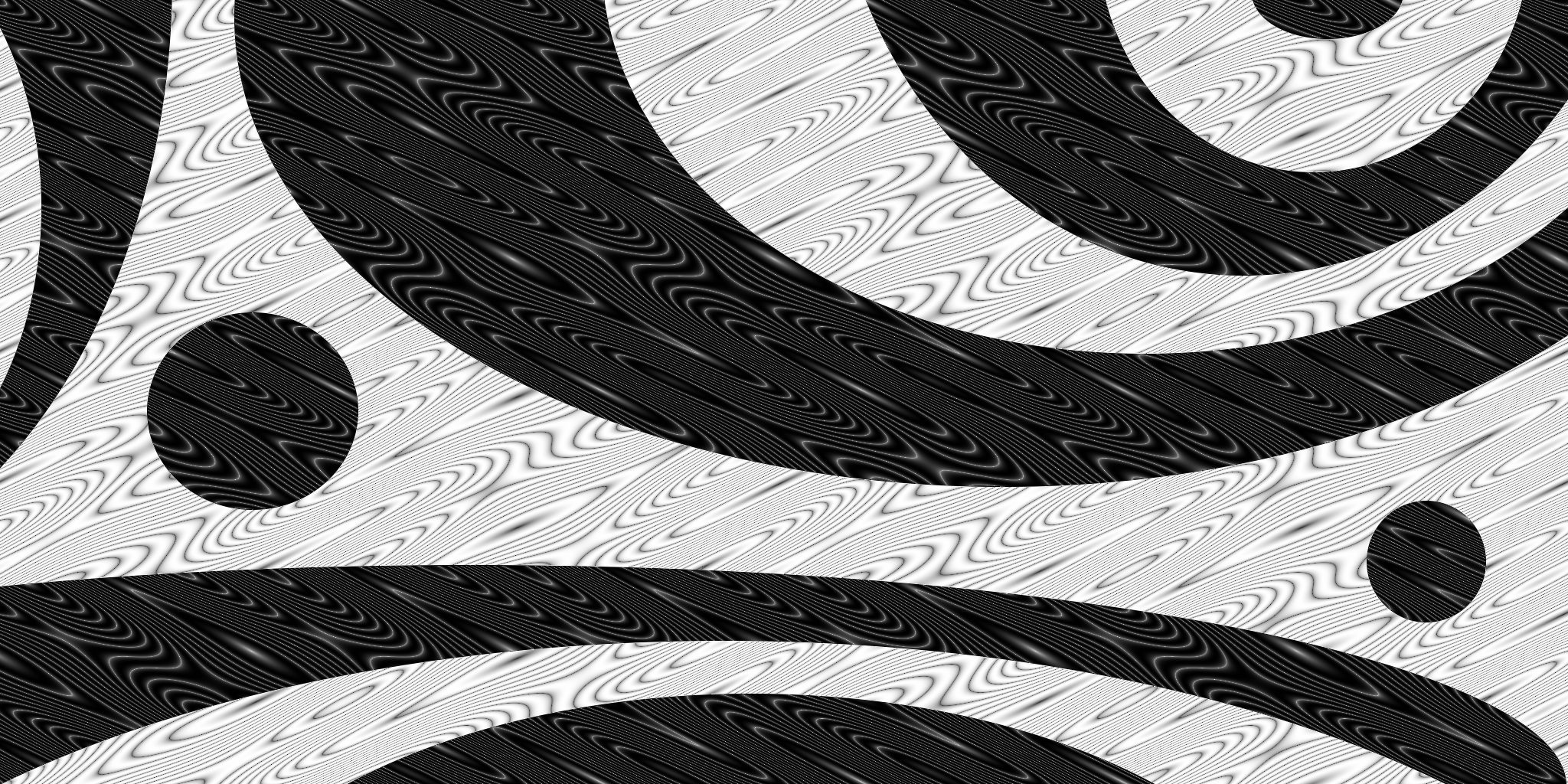}
\caption{``Synthetic'' - Complex patterns.}
\label{synthetic}
\end{figure*}

The fourth image, ``{\em Synthetic}'' (Fig.\,\ref{synthetic}), 
is designed to reveal artifacts that may
be generated by an algorithm when applied to a complex texture such
as sand, grass, fur, cloth, etc., or to a complex repeating pattern when
viewed under perspective transformations or physical deformation (e.g.,
a quilted blanket draped over a person), within which random variations 
may produce coherent structures at different length 
scales\footnote{It must be noted that {\em Synthetic} is extremely
sensitive and will tend to show artifacts such as spurious moire patterns
if scaled, e.g., for this paper.}. Subtle 
directional biases introduced by an algorithm may visibly alter or spuriously
introduce such structures, and this image is designed to be sensitive to
such biases even though it is not ``texture-like'' in appearance.

As can be seen in Fig.\,\ref{synthetic},
{\em Synthetic} has a repeating pattern of radially-parallel structures that
are at the resolution and intensity-discretization limit, i.e., visible artifacts
are unavoidable under spatial and/or intensity decimation. As such it is
sensitive to a class of artifacts that largely subsumes those that are
likely to be revealed by the previous three images. This is a limitation in that
it may reveal artifacts for which the source may be difficult to identify
in the algorithm under examination. On the other hand, its complexity of
structural detail provides sensitivity to artifacts with multivariate
dependencies that may not be revealed by the other three images. 

The next section provides illustrative examples involving use of the
proposed image set.

\section{Examples}

There are clearly too many kinds of image processing operations
and algorithms to permit any sort of comprehensive examination
using the proposed image set. However, ISR and  image compression
represent good candidates for consideration given their
widespread use and the familiar and well-studied artifacts 
produced by various algorithms. To further narrow the scope 
of consideration, examples involving image compression are
restricted to only two generic approaches: decimation-based
local compression, e.g., JPEG~\cite{JPEG}, and spectral-based global 
compression, e.g., using information based on only the
largest singular values from a singular-value
decomposition (SVD) of the image~\cite{SVD}.

\begin{figure}
   \centering
   \includegraphics[width=0.65\linewidth,keepaspectratio]{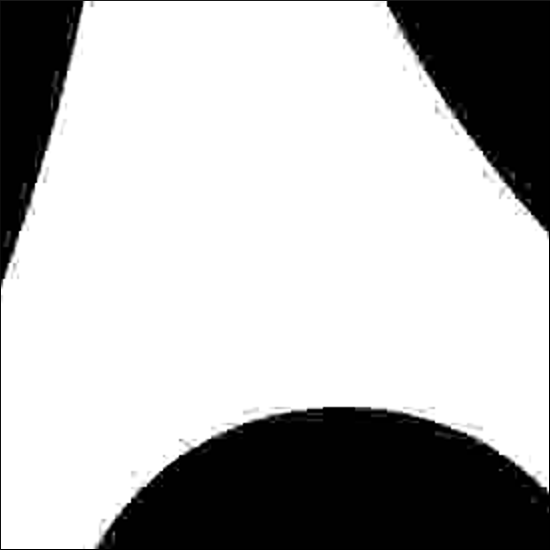}\\ \vspace{3pt}
   \includegraphics[width=0.65\linewidth,keepaspectratio]{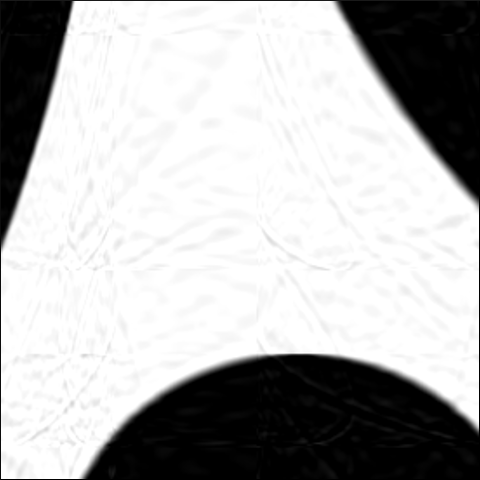}
\caption{Characteristic artifacts from JPEG (Top) and SVD (Bottom) associated
with very high compression in a subregion of {\em Basic} image.}
\label{jpgsvd}
\end{figure}

Typical artifacts of JPEG and SVD in the high-compression
regime\footnote{Because there is no standardized codec
for SVD compression (e.g., with defined bit-depth optimization
of floating-point numbers), examples of SVD and JPEG applied
to the same image in this section do not necessarily 
correspond to equal optimally-achievable compression ratios.} 
are exhibited in the 
subregion of {\em Basic} shown in Fig.\,\ref{jpgsvd}.

The subregion of {\em Basic} in Fig.\,\ref{jpgsvd} shows that JPEG
produces artifacts localized around edges while SVD tends
to preserve smooth edges but introduces artifacts 
globally throughout the image. The JPEG-produced artifacts
around the edges are sometimes referred to as ``chin
whiskers'' in still images or ``mosquito noise'' in video~\cite{ITU}, 
which shows they
have distinctive and recognizable characteristics. The
SVD-produced artifacts also have recognizable characteristics
but are more uniformly distributed, i.e., more noise-like.

\begin{figure}
   \centering
   \includegraphics[width=0.65\linewidth,keepaspectratio]{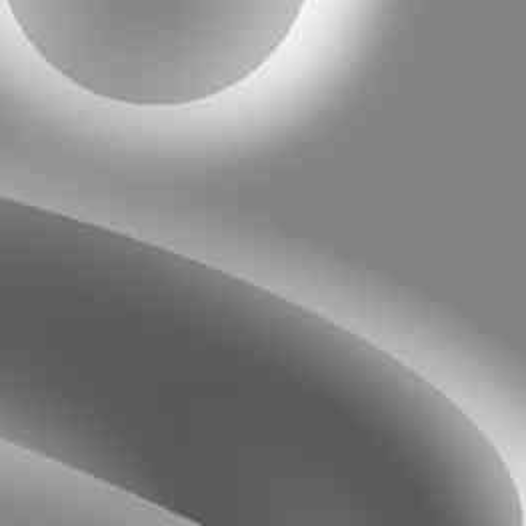}\\ \vspace{3pt}
   \includegraphics[width=0.65\linewidth,keepaspectratio]{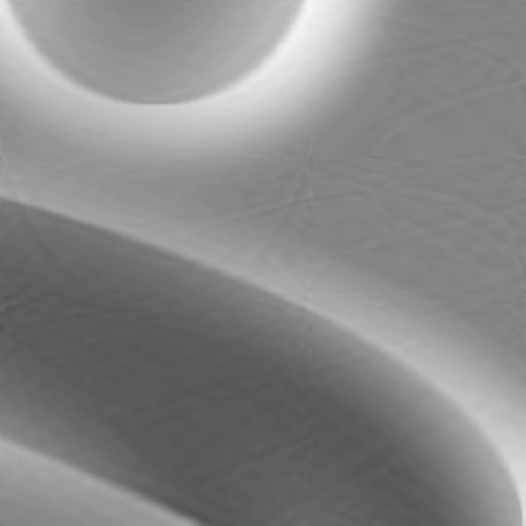}
\caption{High-compression results from JPEG (Top) and SVD (Bottom) 
in a subregion of {\em Platonic} image.}
\label{hcjpgsvd}
\end{figure}

Fig.\,\ref{hcjpgsvd} shows JPEG and SVD high-compression artifacts 
in the lower-right-hand corner of {\em Platonic}. Again, the two
algorithms produce very different, highly distinctive artifacts. 

\begin{figure}
   \centering
   \includegraphics[width=0.65\linewidth,keepaspectratio]{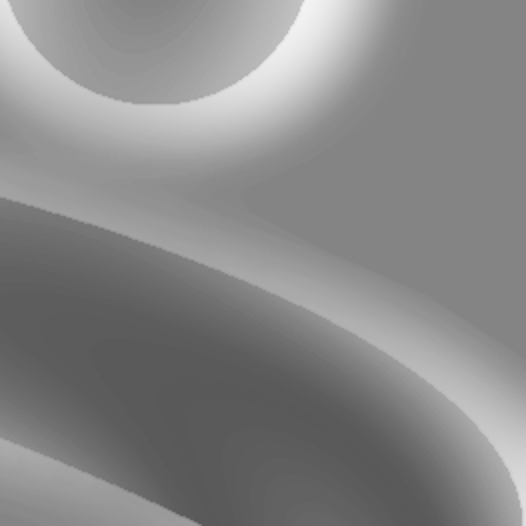}\vspace{3pt}
   \includegraphics[width=0.65\linewidth,keepaspectratio]{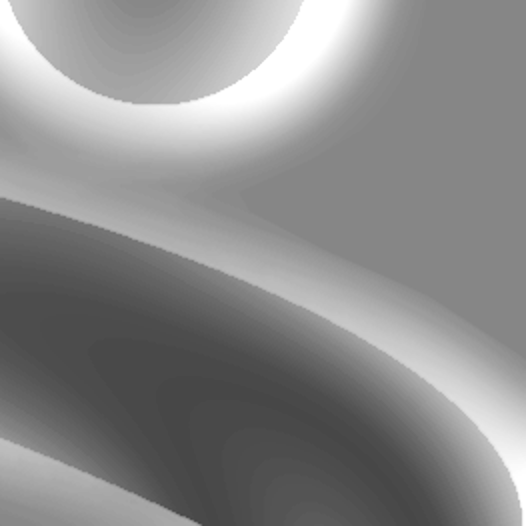}
\caption{Detail of {\em Platonic} (Top) and the same
subregion with an increase in contrast (Bottom) showing spurious contour layering. (Note: Jagged
edges are due to magnification of subregion for easier viewing.)}
\label{contrast}
\end{figure}

\begin{figure}
   \centering
   \includegraphics[width=0.65\linewidth,keepaspectratio]{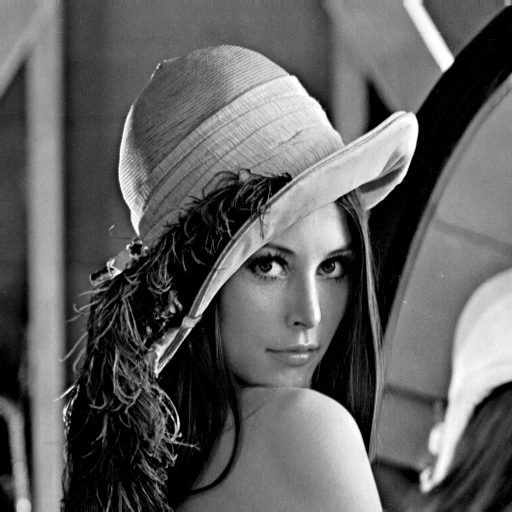}\\ \vspace{3pt}
   \includegraphics[width=0.65\linewidth,keepaspectratio]{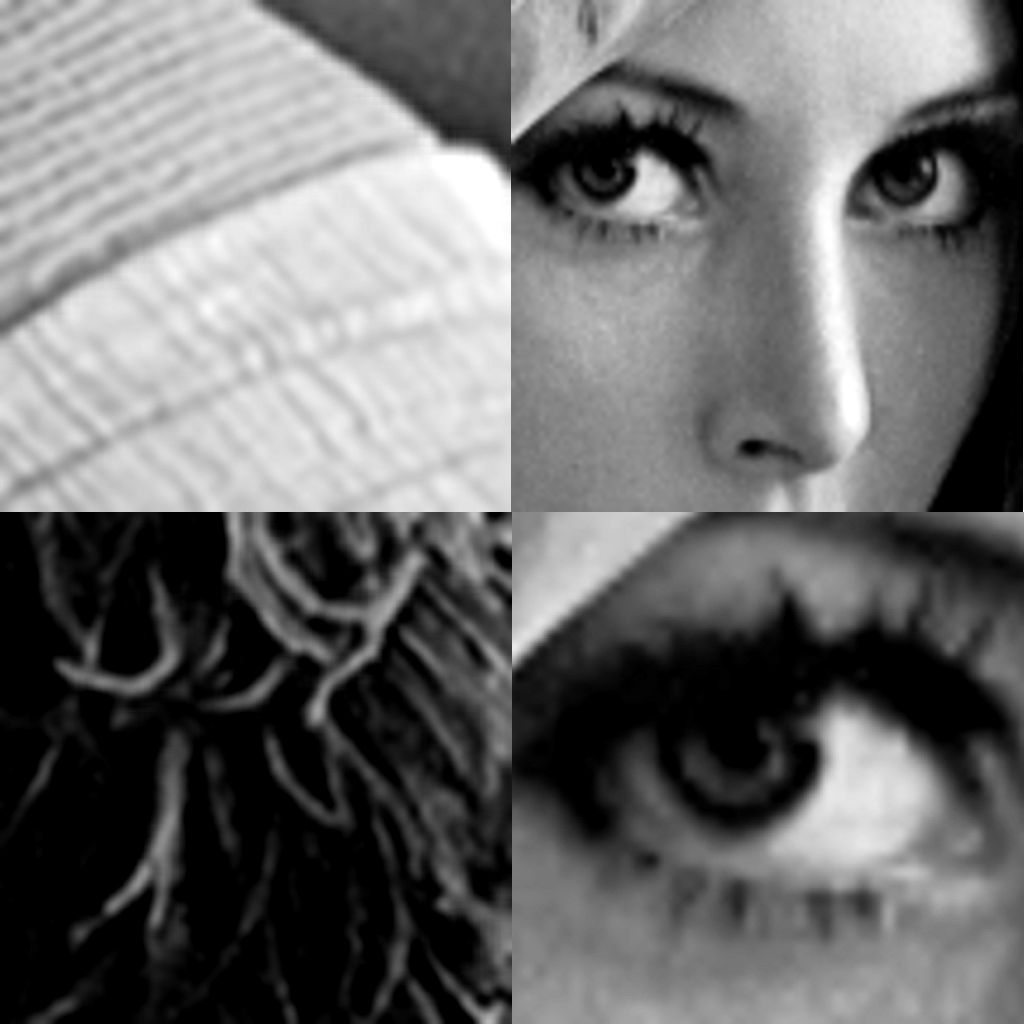}
\caption{Contrast enhancement of {\em Lena} does not reveal any 
clearly-visible artifacts that are not present in the original image, i.e., its sensitivity is low as a test image.}
\label{contrastLena}
\end{figure}

Fig.\,\ref{contrast} shows the effect of increasing contrast\footnote{All examples involving
contrast enhancement were produced by GIMP at level 30.} to a subregion of {\em Platonic}.
Specifically, applying an increase in contrast to {\em Platonic} produces
significant banding in regions of smoothly-varying intensity and saturation artifacts in
the corona at the center-top of the image.  Fig.\,\ref{contrastLena} shows
the same increase in contrast applied to {\em Lena}, resulting in no clearly-visible 
generation of artifacts. 

\begin{figure}
   \centering
   \includegraphics[width=0.65\linewidth,keepaspectratio]{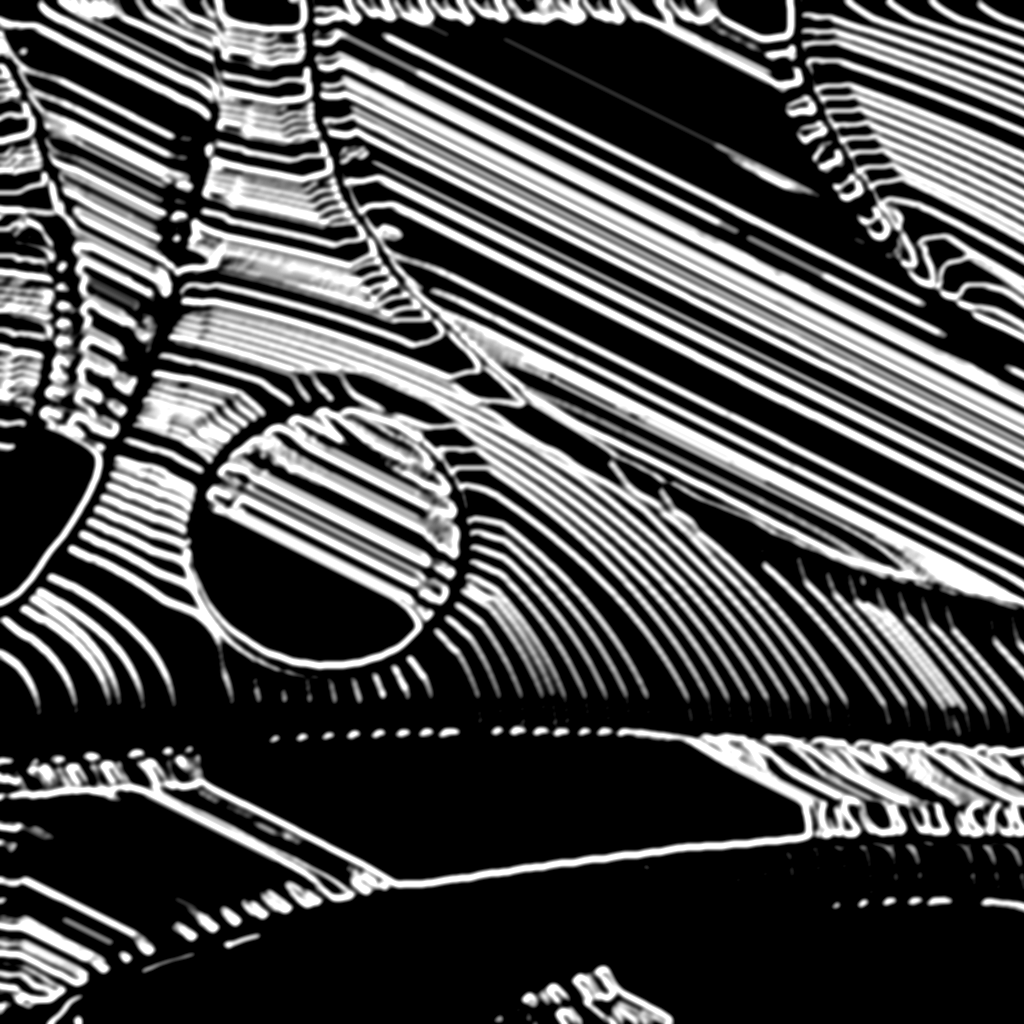}\\ \vspace{3pt}
   \includegraphics[width=0.65\linewidth,keepaspectratio]{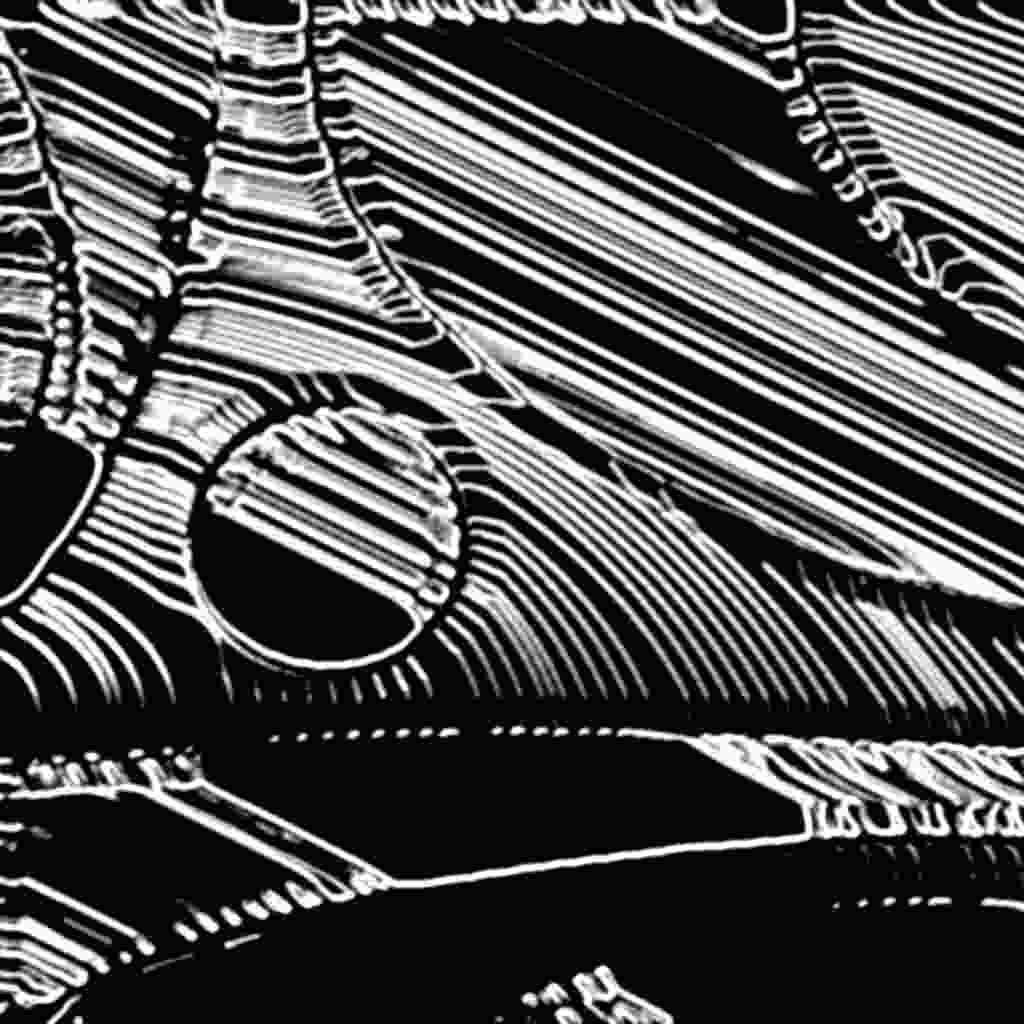}
\caption{Left half of {\em Natural} image (Top) and high-compression 
results from JPEG (Bottom) with visible pixelation noise and aliasing.}
\label{natcomp}
\end{figure}

Fig.\,\ref{natcomp} shows high-compression artifacts from JPEG when applied to {\em Natural}. 

\begin{figure}
   \centering
   \includegraphics[width=0.65\linewidth,keepaspectratio]{Nat-half-orig.png}\\ \vspace{3pt}
   \includegraphics[width=0.65\linewidth,keepaspectratio]{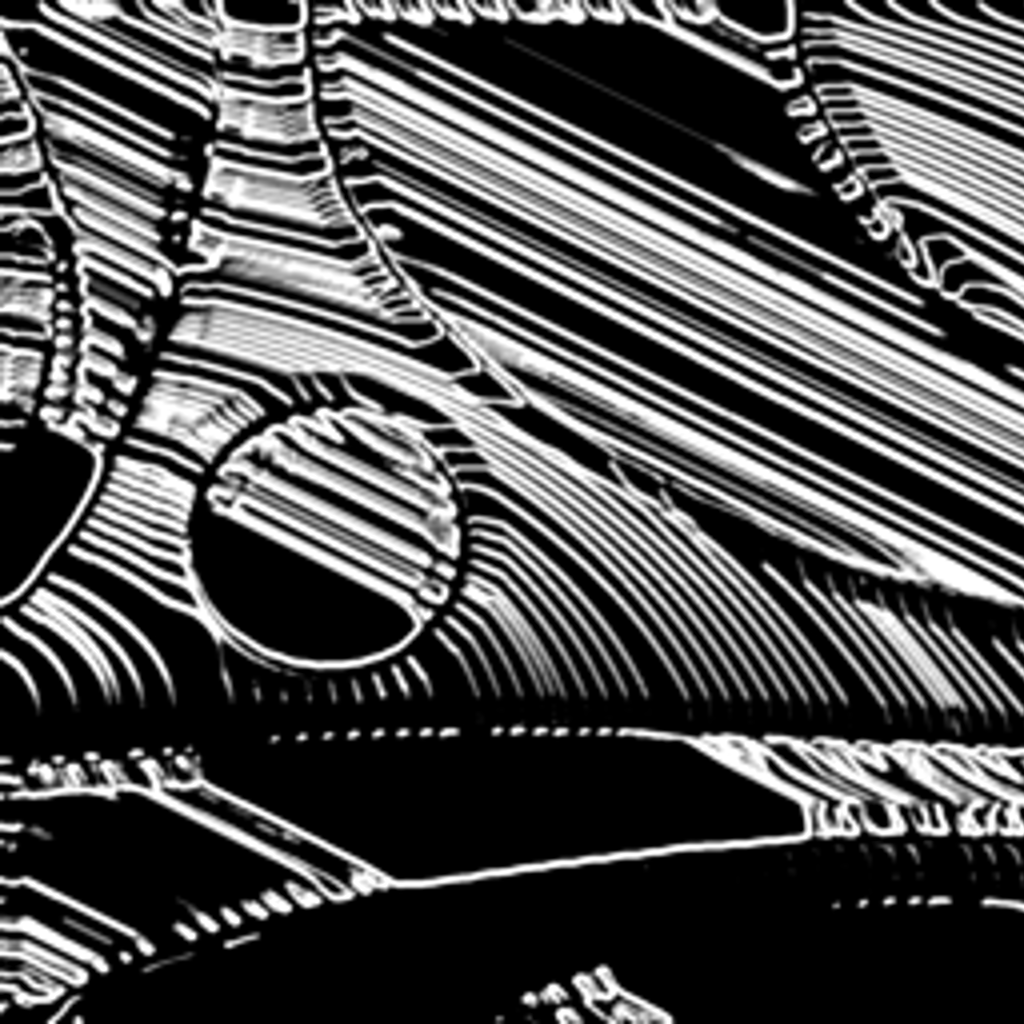}
\caption{Left half of {\em Natural} image (Top) and the result of a 4x down-res 
followed by a 4x up-res (Bottom) with clearly-visible pixelation and aliasing artifacts.}
\label{natres}
\end{figure}

Fig.\,\ref{natres} shows results from the application of a cubic 4x axis decimation (down-res)
of {\em Natural} followed by cubic up-res (super-resolution/interpolation) back to its 
original dimensions. 

\begin{figure}
   \centering
   \includegraphics[width=0.65\linewidth,keepaspectratio]{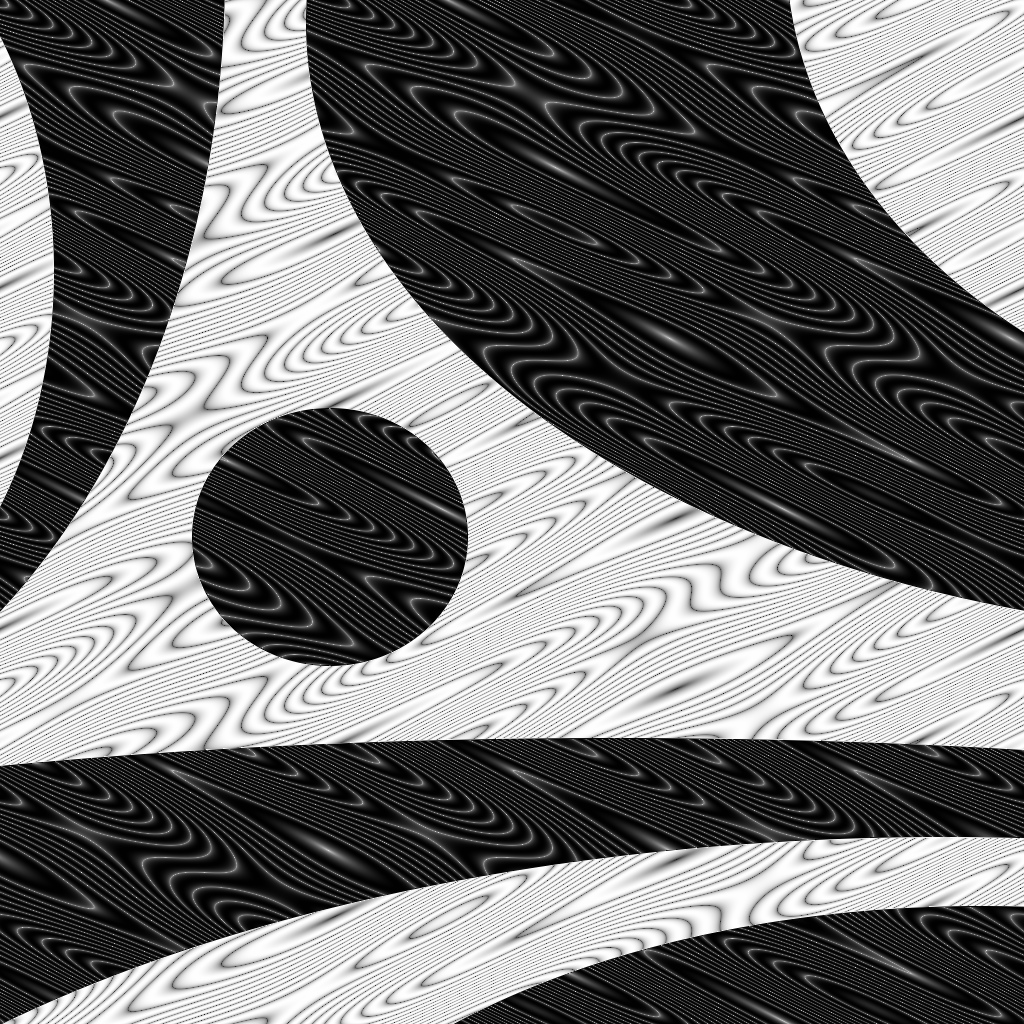}\\ \vspace{3pt}
   \includegraphics[width=0.65\linewidth,keepaspectratio]{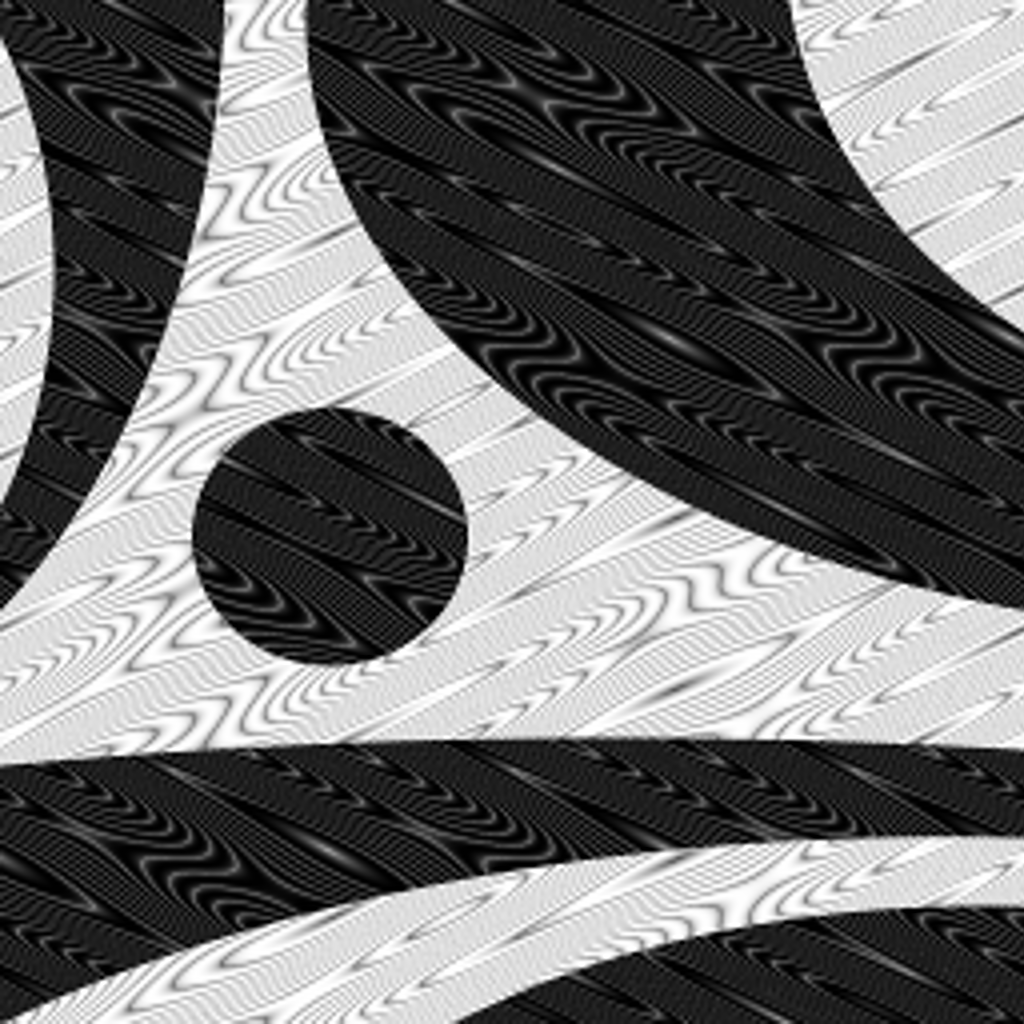}
\caption{Left half of {\em Synthetic} image (Top) and the result of a 4x down-res 
followed by a 4x up-res (Bottom) with clearly-visible pixelation and aliasing artifacts.}
\label{synres}
\end{figure}

Fig.\,\ref{synres} shows the same operation as Fig.\,\ref{natres} but applied to {\em Synthetic}.

\begin{figure*}
   \centering
   \includegraphics[width=1.0\textwidth,keepaspectratio]{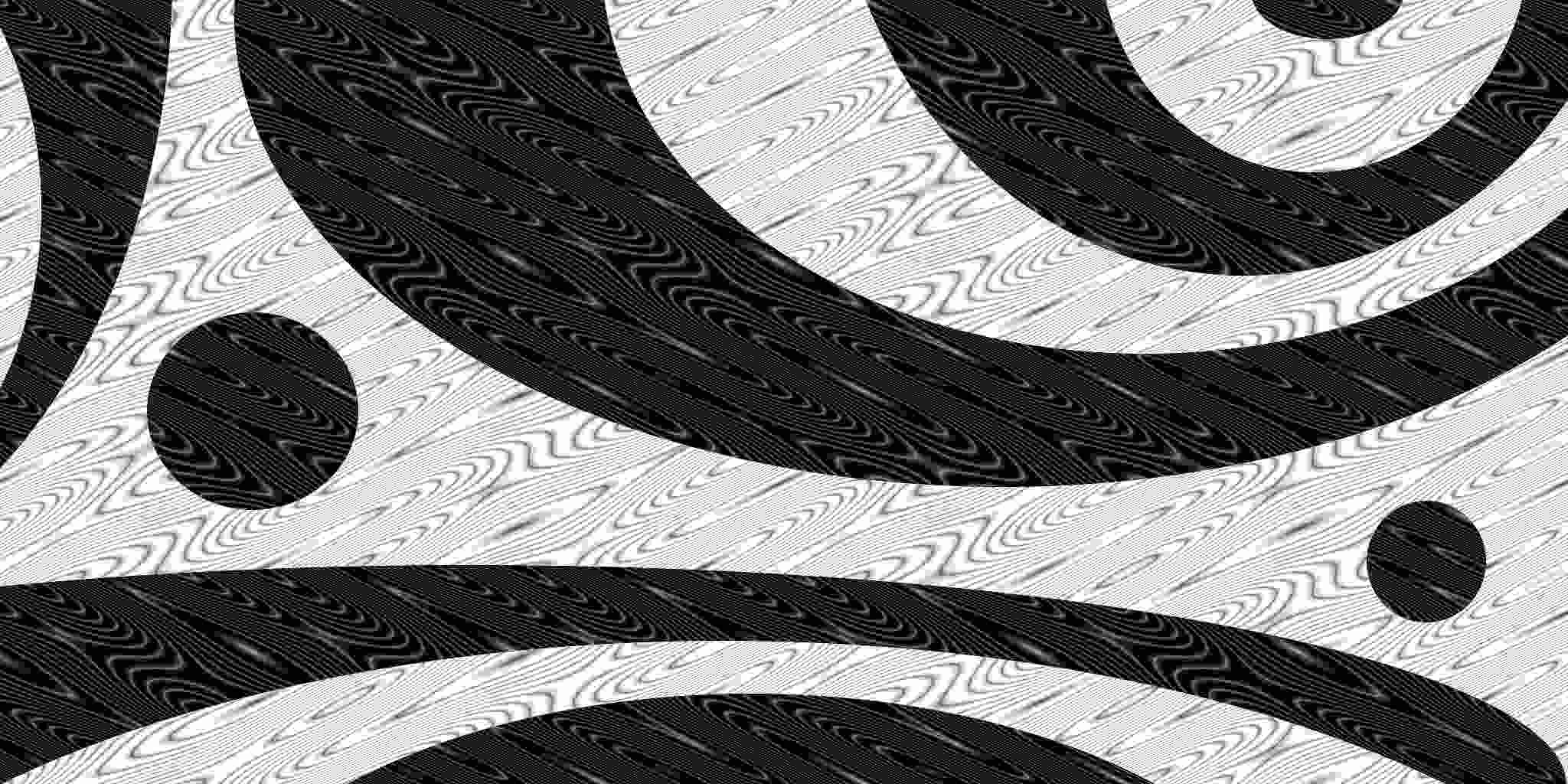}
\caption{JPEG high-compression of {\em Synthetic}.}
\label{synjpg}
\end{figure*}

\begin{figure*}
   \centering
   \includegraphics[width=1.0\linewidth,keepaspectratio]{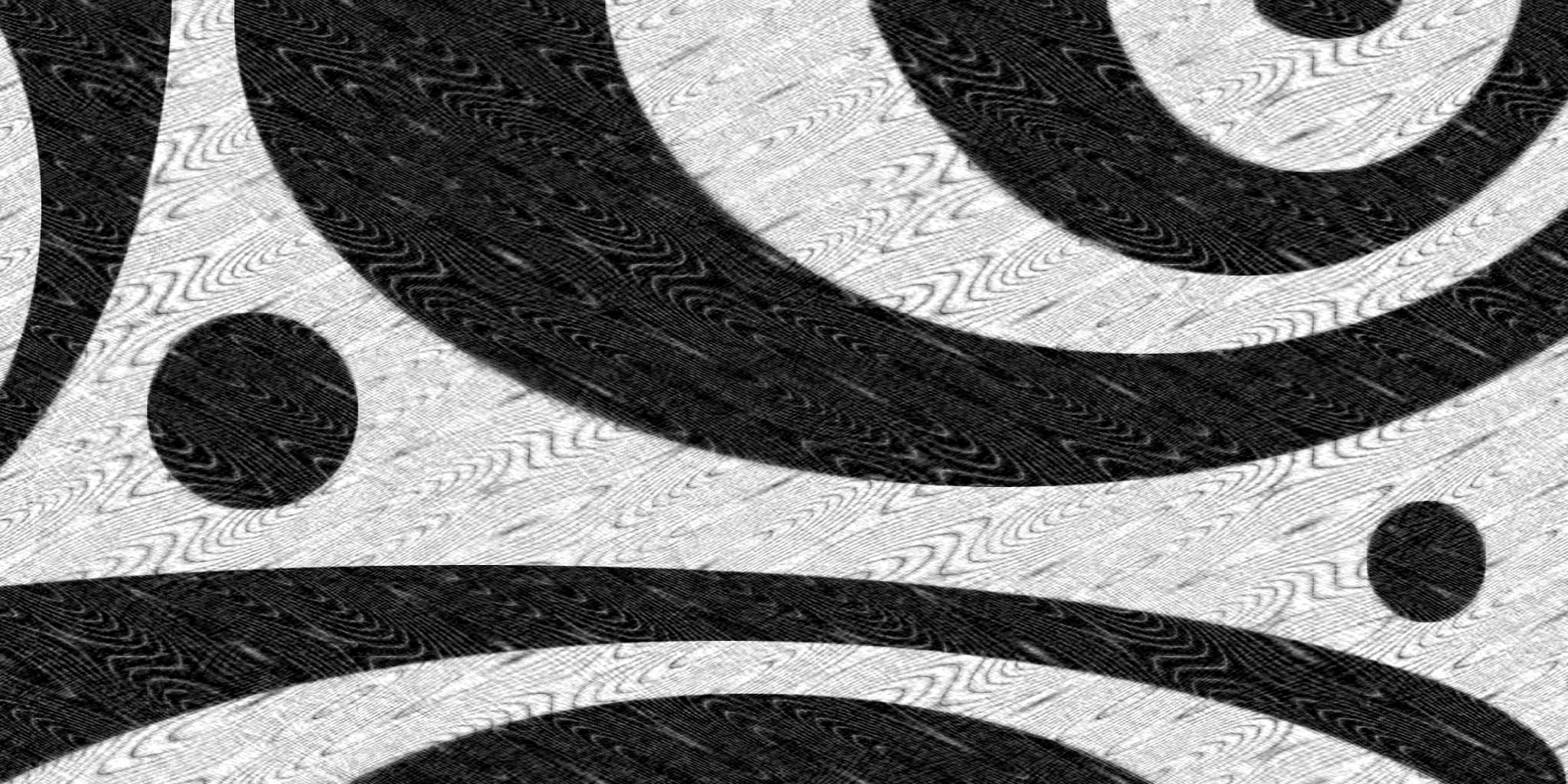}
\caption{SVD high-compression of {\em Synthetic}.}
\label{synsvd}
\end{figure*}

Lastly, Fig.\,\ref{synjpg} and Fig.\,\ref{synsvd} respectively show high-compression results from applying
JPEG and SVD to {\em Synthetic}.
In both cases the distinctive artifacts associated with JPEG and SVD 
compression are clearly visible. In the case of JPEG the artifacts
are manifest in multiple ways, including moire patterns, while the
SVD artifacts are of a form that resembles scratches and surface
dust.

\section{Discussion}

Four artificially-generated images have been proposed as candidate test images
for assessing and comparing the qualitative behaviors of different image 
processing algorithms. These images have been examined in a wide variety of
contexts since they were developed at the Naval Research Laboratory (NRL)
in the 1990s. There is no rigorous and general statement that can be made
except that the four images have been found to be distinct in their specificity to 
different types of artifact. It is clearly also the case that they do not contain
(by design) interpretable content that can lead to subjective bias when 
applied to qualitatively assess the properties of a given algorithm. 
Anecdotal experience motivates significantly stronger claims, but here we
only put forth this set of  images as candidates for consideration as
potential standards for the image processing community.

\end{document}